\pdfoutput=1
\documentclass[11pt,a4paper]{article}
\usepackage[hyperref]{emnlp2020}
\usepackage{times}
\usepackage{latexsym}
\usepackage{subcaption}
\usepackage{booktabs}
\usepackage{amssymb}
\usepackage{amsmath}
\usepackage{comment}
\usepackage{url}
\usepackage{enumitem}
\usepackage{tikz}
\usepackage{array}
\usepackage{xspace}
\usepackage{hyperref}
\usepackage{todonotes}

\usepackage[nameinlink]{cleveref}
\crefformat{section}{\S#2#1#3} 
\crefformat{subsection}{\S#2#1#3}
\crefformat{subsubsection}{\S#2#1#3}

\usepackage{amsmath,amsfonts,bm}
\usepackage{mathtools}

\def\eqref#1{equation~\ref{#1}}

\def\1{\bm{1}}

\def\vc{{\bm{c}}}

\DeclareMathAlphabet{\mathsfit}{\encodingdefault}{\sfdefault}{m}{sl}
\SetMathAlphabet{\mathsfit}{bold}{\encodingdefault}{\sfdefault}{bx}{n}

\def\gD{{\mathcal{D}}}

\def\gL{{\mathcal{L}}}
\def\gM{{\mathcal{M}}}

\def\gR{{\mathcal{R}}}
\def\gS{{\mathcal{S}}}
\def\gT{{\mathcal{T}}}

\newcommand{\Ls}{\mathcal{L}}

\usepackage{microtype}

\aclfinalcopy 
\def\aclpaperid{3227} 

\newcommand{\red}[1]{\textcolor{red}{#1}}
\newcommand{\blue}[1]{\textcolor{blue}{#1}}

\newcommand{\ie}{{\em i.e.,}\xspace}
\newcommand{\eg}{{\em e.g.,}\xspace}

\newcommand{\Ni}{({\em i})~}
\newcommand{\Nii}{({\em ii})~}

\newcommand{\concat}{{\sc Concat}}
\newcommand{\sts}{{\sc Sen2Sen}}

\DeclareMathAlphabet{\pazocal}{OMS}{zplm}{m}{n}
\DeclareMathAlphabet{\pazocal}{OMS}{zplm}{m}{n}

\title{Pronoun-Targeted Fine-tuning for NMT with Hybrid Losses}

\author{Prathyusha Jwalapuram$^*$, Shafiq Joty$^*$$^\S$, Youlin Shen$^*$\\
  $^*$Nanyang Technological University, Singapore \\
  $^\S$Salesforce Research Asia, Singapore \\
$^*${\tt\{jwal0001,srjoty,yshen010\}@ntu.edu.sg} \\
\\}

\date{}

\begin{document}
\maketitle
\begin{abstract}

Popular Neural Machine Translation model training uses strategies like backtranslation to improve BLEU scores, requiring large amounts of additional data and training. We introduce a class of conditional generative-discriminative hybrid losses that we use to fine-tune a trained machine translation model. Through a combination of targeted fine-tuning objectives and intuitive re-use of the training data the model has failed to adequately learn from, we improve the model performance of both a sentence-level and a contextual model without using any additional data. We target the improvement of pronoun translations through our fine-tuning and evaluate our models on a pronoun benchmark testset. Our sentence-level model shows a 0.5 BLEU improvement on both the WMT14 and the IWSLT13 De-En testsets, while our contextual model achieves the best results, improving from 31.81 to 32 BLEU on WMT14 De-En testset, and from 32.10 to 33.13 on the IWSLT13 De-En testset, with corresponding improvements in pronoun translation. We further show the generalizability of our method by reproducing the improvements on two additional language pairs, Fr-En and Cs-En.\footnote{Code available at \url{<https://github.com/ntunlp/pronoun-finetuning>}.}

\end{abstract}

\section{Introduction}
\label{sec:intro}
{The advent of neural machine translation (NMT) \cite{Bahdanau2015NeuralMT, Vaswani-17-transformer} brought about significant improvements that left the previously successful statistical machine translation models far behind. However, the availability of large corpora has been no small part of that success, with recent NMT models using millions of sentences for training. A lack of availability of such large parallel corpora across languages has given rise to methods utilizing large amounts of monolingual data, such as for backtranslation \cite{sennrich-etal-2016-improving}, language modeling \cite{Glehre2017LanguageModelNMT,mirror_Zheng2020Mirror-Generative}, or for large-scale pre-training \cite{Lewis2019BARTDS}.}

Backtranslation \cite{sennrich-etal-2016-improving,EdunovBacktranslationatscale} is a commonly used strategy to improve MT models in the absence of adequate parallel data for training. A target-to-source model is first trained using the available parallel data, which is then used to translate a large target-monolingual corpus into the source to create pseudo-parallel data for training a source-to-target MT model. This has been shown to result in improvements in the BLEU score, and has become a popular method for improving NMT models, with many recent works proposing strategies to further improve it \cite{Hoang2018IterativeBacktranslation, Yang2019EffectivelyTN, Caswell2019TaggedB}.  However, recent studies have suggested that there is a limit beyond which the addition of synthetic data hurts the performance of the model \cite{Fadaee2018BackTranslationSB, Poncelas2018InvestigatingBI}.  Also, recent work \citep{eval_back_translation_translationese,nguyen2019data} point out that back-translation suffers from the \emph{translationese effect}, where back-translation only improves the performance when the source sentences are translationese but does not offer any improvement when the sentences are natural text.

Automatic post-editing (APE) is another common strategy that is used to improve translations. APE models are commonly monolingual, and typically take the output from some MT model as input, which they then modify. In the absence of adequate human post-edited data to train data-hungry neural models, \citet{Voita2019ContextAwareMR} and \citet{Freitag2019TextRM} both use round-trip translation data to train their post-editing models. {In round-trip translations, target monolingual data is translated using a target-to-source model to the source text, and then back to the target using another source-to-target model. This round-trip translated text is considered an approximation of poor quality MT output, which can be used in combination with the original target reference text to train the post-editing model. \citet{Voita2019ContextAwareMR} train a model to make corrections in context, using groups of 4 sentences as input, and show improvements in BLEU as well as translations of discourse phenomena.}

NMT models typically fail on rare words that may not be adequately seen during training, such as named entities, or on words whose interpretation depends on the context such as discourse phenomena  \cite{koehn-knowles-2017-six, SennrichDisourceMT}. 
{For the latter, NMT models tend to prefer a more typical alternative to a relatively rare but correct one {(\eg\ French ``\textit{Il}" is often wrongly translated to the more common ``\textit{it}" than ``\textit{he}" )}. However, these seemingly trivial errors can erode translation to the extent that they can be easily distinguishable from human-translated texts \cite{Lubli2018HasMT}.} 

There could be several reasons for why NMT models make such mistakes; our hypothesis is that since almost all NMT models are trained with a conditional language model objective, it is clear that this objective alone is proving inadequate to capture all of the information available in the text. We therefore propose a class of conditional generative-discriminative hybrid losses that explicitly teach models what to generate and what not to generate. Using these specialized losses, we aim to improve the learning power of the MT model.




{Specifically, in this work, we target the improvement of pronoun translation by focusing our fine-tuning efforts through our proposed objectives and also through the fine-tuning data. We aim to leverage the training data we already have by extracting a subset of targeted fine-tuning data from the training corpus that the model has failed to learn correctly from.  We use the newly proposed training objectives in combination with the targeted data to help the model fully reach its learning potential on the training corpus.  We attempt to improve both general translation quality and the pronoun translation without compromising on either, and to do this without any elaborate model architecture. }

Our main contributions are as follows:

\begin{itemize}[leftmargin=*, itemsep=0.3pt, topsep=-0.0pt]
    \item A class of Conditional Generative-Discriminative Hybrid losses that improve the learning potential of the model (\cref{sec:method:obj}).
    \item Effective fine-tuning strategy that uses the training data itself to improve MT (\cref{sec:finetune}).
    \item Improvements in BLEU over WMT14 and IWSLT13 De-En testsets, and in pronoun translations over a pronoun challenge testset (\cref{subsection:loss_finetuning}, \cref{subsec:iwslt}).
    \item {Demonstration of generalizability through additional fine-tuning experiments on Fr-En and Cs-En (\cref{subsec:other_langs}).}
\end{itemize}




\section{Targeted Finetuning Objectives} \label{sec:method:obj}

Before introducing our proposed Conditional Generative-Discriminative hybrid losses for fine-tuning NMT models on a targeted dataset, we first describe the Conditional Language Modeling (CLM) objective used to train NMT models.

\subsection{Conditional Language Modeling} 
NMT models are generally trained with the CLM generative loss that relies on an auto-regressive factorization to perform density estimation and generation of target texts. For a source-target sentence pair $(x,y)$, a CLM predicts a conditional probability distribution $P_{\theta}(y_{1:n}|x)$, where $n$ is the number of tokens in the target text. The auto-regressive factorization for a CLM is given by
\begin{equation}
P_{\theta}(y_{1:n}|x) = \prod_{t=1}^{n} P_{\theta}(y_t|y_{<t}, \vc) \label{eq:clm}
\vspace{-0.2em}
\end{equation} 
where $\vc$ is a context vector that summarizes the relevant input (\eg\ attended vector over source text and the current decoder state). The CLM training objective for NMT can be written as: 
\begin{equation}
\label{eq:genloss}
\gL_g  =  - \frac{1}{n} \sum_{t=1}^{n} \log P_{\theta} (y_t|y_{<t}, \vc)
\vspace{-0.2em}
\end{equation}
Generating from CLM trained NMT models requires iteratively sampling from $P_{\theta} (y_t|y_{<t}, \vc)$, and then feeding $y_t$ back into the model as input.

\subsection{Generative-Discriminative Hybrid Loss}


While CLM has been the de-facto loss to train NMT models, models trained with CLM make mistakes that can erode translation quality, making them easily distinguishable from human translation. For example, state-of-the-art NMT models are not very good at handling rare words like named entities. They have also been criticized for not being sensitive to discourse-level aspects such as pronouns, lexical consistency, and discourse connectives \cite{SennrichDisourceMT,Jwalapuram2020CanYC}.

We introduce a generative-discriminative hybrid method for fine-tuning NMT models, with the motivation of generating tokens that are more strongly in one class vs. another. We consider that the reference tokens come from a positive class, whereas the model generated tokens come from a negative class. We propose two variants of our hybrid training -- \Ni log-likelihood and \Nii max-margin.


\paragraph{Log-likelihood training.}

Let $z \in \{0,1\}$ represent the class for a training instance $(x, y)$. We can consider a generative classifier as follows. 
\begin{equation}
\small
P_{\theta}(z = k| x,y)  =  \frac{P (z = k)   P_{\theta}(y|x, z=k)}{\sum_{k'=0}^1 P (z = k^{'})  P_{\theta}(y|x,z= k^{'} )} \label{eq:gencls}
\end{equation}

\noindent Assuming an equal prior class probability, \ie\ $P (z = 1) = P (z = 0)$ and by replacing $P_{\theta}(y|x, z=k)$ with \Cref{eq:clm}, we can write: 
\begin{equation}
\small 
P_{\theta}(z=k| x,y)  =  \frac{\prod_{t=1}^{n} P_{\theta}(y_t|y_{<t}, \vc, k)}{\sum_{k^{'}} \prod_{t=1}^{n} P_{\theta}(y_t|y_{<t}, \vc, k^{'})}\label{eq:gencls2}
\end{equation}

\noindent Since our objective is to maximize the probability of the reference tokens, we minimize the following negative log-likelihood loss:
\begin{equation}
\small
\gL_{nll} =  - \log P_{\theta}(z=1| x,y) 
\end{equation}

If $y^{+}$ is the reference (positive) translation and $y^{-}$ is the model (negative) output, it is easy to show that the above loss is equivalent to  
\begin{equation}
\small 
{\Ls_{nll} =  - \frac{1}{n} \sum_{t=1}^{n} \log  {\frac{\exp( \hat{y}^{+}_t /\tau) }{ \Big( \exp( \hat{y}^{+}_t/\tau) + \exp( \hat{y}^{-}_{t}/\tau) \Big) }}} \label{eq:nll}
\end{equation}
where $\tau$ is the temperature parameter of the softmax,\footnote{For the sake of simplicity, we omit $\tau$ in Eq. \ref{eq:gencls} - \ref{eq:gencls2}} and $\hat{y}^{+}_t$ and $\hat{y}^{-}_{t}$ are the final-layer logits (pre-softmax activations) corresponding to the reference token $y_t^{+}$ and model generated token $y_t^{-}$, respectively. The logit for the model generated token is computed by just taking the $\max$ over all the logits. We use $\tau=0.5$ for our experiments.





\paragraph{Max-margin training.}

Following \citet{collobert2011natural}, we also propose a pairwise ranking loss that maximizes the distance between positive and negative samples. Formally, 
\begin{equation}
\Ls_{mm} =  \frac{1}{n} \sum_{t=1}^{n} \max \{0, \mu - \hat{y}^{+}_t + \hat{y}^{-}_{t} \} \label{eq:mm}
\end{equation}
where $\mu$ is the margin; we use $\mu=0.3$. 



{Note that the additional losses can be applied to all the tokens in the sequence, or restricted to some tokens. We demonstrate this in our experiments by applying the loss on all tokens and  selectively applying the loss only on pronouns.}
Both of the discriminative losses essentially promote the probability of the positive (\ie correct) sample. However, the intuition behind using the additional loss over the standard loss is that the fine-tuning here focuses on improving the positive sample over the negative sample that the model has learnt to produce, rather than over the entire probability distribution over the full vocabulary. 

We average these losses at both the sentence and the batch-level to add it to the existing CLM  loss. The overall loss for training is 
\begin{equation}
\label{eq:hybrid_obj}
\gL_{gd} = \lambda \gL_g + (1-\lambda) \gL_d
\end{equation}
where $\lambda$ is a weighting hyperparameter, and the discriminative loss $\gL_d$ is either $\gL_{mm}$ (Eq. \ref{eq:mm}) or $\gL_{nll}$ (Eq. \ref{eq:nll}). In our training, the discriminative loss $\mathcal{L}_d$ is aimed at correcting the mistakes, whereas the generative loss $\mathcal{L}_g$ is needed to preserve the translation adequacy and fluency. In our experiments, we simply set $\lambda = 0.5$.




\section{Fine-tuning Data \& MT Baselines} \label{sec:finetune}


\subsection{Pronoun-Targeted Fine-tuning Data}
\label{subsec:method:finetune_data}
{We create a subset of the training corpus in order to find training data that has not been fully learnt from; particularly, we focus our fine-tuning experiments on pronoun translation.} {Pronouns are an important discourse phenomenon that provide references to entities that have previously occurred in a text. Mistranslations can lead to loss of grammaticality or inference of the wrong antecedent, resulting in a misunderstanding of the text \cite{Guillou2012ImprovingPT}.   }



{Consider a parallel corpus $\gD = (\gS, \gR)$, where $\gS$ is the source and $\gR$ is the target/reference text. Assuming that the baseline NMT models (\cref{sec:method:baselines}) are trained {until convergence} using this data, for our targeted fine-tuning of pronoun translations, we derive a subset of the training corpus $\gD$ as follows:}

\begin{enumerate}[leftmargin=*,itemsep=0.1pt]
    \item Translate $\gD$ using a baseline model $\gM$ to obtain source to target translations $\gT_{\gM}$.
    \item Align $\gT_{\gM}$ with reference $\gR$ {using} \textit{efmaral} \cite{ostling2016efmaral}.
    \item Find pronoun translations in $\gT_{\gM}$ that do not match reference $\gR$. To exclude equivalent but non-identical translations, we use the list provided by \citet{EvalPronom}\footnote{https://github.com/ntunlp/eval-anaphora}.
    \item For each sentence with a mistranslated pronoun, extract the source sentences from $\gS$. 
    \item The corresponding source and reference sentences form the pronoun-targeted fine-tuning subset, referred to as $\gD_{\text{prn}} = (\gS^{'}, \gT^{'})$. 
\end{enumerate}


\subsection{Baseline MT Models} \label{sec:method:baselines}

{Typically, MT models are trained at the sentence level, taking one sentence as input and producing one sentence as output. Most MT systems at the sentence-level do not have access to adequate context that may be required for the translation of pronouns \cite{SennrichDisourceMT}. Since it is our aim to improve pronoun translation, we train both a sentence-level model and a simple concatenation-based contextual model as our baselines:}

\begin{description}[leftmargin=*,itemsep=0.3pt]
\item [\sts: ] {A standard 6-layer base Transformer model \cite{Vaswani-17-transformer} trained to translate each sentence independently.} 

\item [\concat: ] {A standard 6-layer base Transformer trained to translate a sentence given one previous sentence as context \cite{tiedemann-scherrer-2017-neural}. The input to the model is the previous sentence and the current sentence combined with a special separator character. {\citet{Jwalapuram2020CanYC} show that this simple context model performs comparably or better than other elaborate contextual models like \citet{Voita-ACL18}, \citet{zhang-etal-2018-improving}, and \citet{ miculicich-etal-2018-document}.}} 
\end{description}

\noindent Both the baseline models are trained for 100,000 steps. Other parameter details are in the Appendix.

\section{Experiments}
\label{sec:experiments}
{We conduct our fine-tuning experiments on the German-English (De-En) translation task. We describe our baseline training and fine-tuning corpus (\cref{subsection:data}), our experiments and results on fine-tuning using only the targeted subset data (\cref{subsection:data_finetuning}), and fine-tuning using both the targeted subset data and the hybrid training losses (\cref{subsection:loss_finetuning}). }

\subsection{MT Training Data}
\label{subsection:data}

\paragraph{Baseline training corpus.} {We use a De-En training dataset consisting of about 2.5 million sentence pairs, taken from the News Commentary, IWSLT \cite{IWSLT} and Europarl \cite{Europarl} corpora.\footnote{We exclude the UN corpus as our analysis showed that it does not have a high incidence of pronouns.} 
Sentences are encoded through Byte-Pair Encoding (BPE) \cite{Sennrich2016BPE} with 40,000 operations, which results in a shared vocabulary of 40,224 tokens. We will refer to our baseline dataset as $\gD$.} 

\vspace{-0.5em}
\paragraph{Pronoun targeted fine-tuning data.} {As described in \cref{subsec:method:finetune_data}, we derive the pronoun-targeted fine-tuning subset $\gD_{\text{prn}}$ from the baseline training corpus $\gD$ based on the translation errors of the baseline models. This results in a pronoun-targeted subset of 294,535 pairs for the \sts\ model and 285,783 pairs for the \concat\ model.}


\vspace{-0.5em}
\paragraph{Random subset.} {We randomly extract a subset of 300,000 sentence pairs from $\gD$, which approximately matches the size of the pronoun-targeted subset. We will refer to this dataset as $\gD_{\text{rand}}$.}


\subsection{Pronoun Translation Evaluation}

\paragraph{Testset.} We run the models on the pronoun challenge testset provided by \citet{EvalPronom}, which is extracted from WMT testsets based on submission errors. For De-En, the testset has 2245 sentences, taken from WMT17-WMT19. 

\vspace{-0.5em}
\paragraph{Evaluation.} {We report the macro-averaged F1 scores of the pronoun translation based on a simplified version of AutoPRF \cite{HardmeierAutoPRF}. For each sentence in the testset, the counts of the pronouns in the system translation are clipped based on the pronouns in the reference translation; these counts are then used to compute the precision, recall and F1 scores. }



\subsection{Baseline Results}
\label{subsection:baselines}

\begin{table}[t]
\centering
\scalebox{0.75}{\begin{tabular}{lc|ccccc}
\toprule
 &  & \textbf{WMT14}  & \multicolumn{4}{c}{\textbf{Pronoun Testset}} \\
 \cmidrule(lr){3-3} \cmidrule(lr){4-7} 
\textbf{Model} & \textbf{Train}  & \textbf{BLEU} & \textbf{BLEU}  &\textbf{P} & \textbf{R} &\textbf{F1} \\ 

\midrule


\sts\ & $\gD$ & 31.64 & 35.56 & 77.92 & 66.01 & 69.55  \\
\concat & $\gD$ & \textbf{31.81} & \textbf{36.16} & \textbf{80.39} & \textbf{68.49} & \textbf{72.03} \\
 
\bottomrule
\end{tabular}}
\caption{Baseline BLEU results on the WMT14 De-En testset and the BLEU (for translation), \textbf{P}recision, \textbf{R}ecall and \textbf{F1} scores (for pronoun translations) on the pronoun testset from \citet{EvalPronom}.}
\label{table:baseline_results}

\end{table}






We first report the BLEU scores on the WMT14 De-En testset, and the BLEU, precision, recall, and F1 scores on the pronoun testset from \citet{EvalPronom} in Table~\ref{table:baseline_results}. The \sts\ model results in a BLEU of \textbf{31.64}, while the \concat\ model results in a slightly higher performance at \textbf{31.81} BLEU. More importantly, there is an improvement in the pronoun translations, with the F1 increasing from 69.55 for the \sts\ model to 72.03 for the \concat\ model. To confirm that the context provides useful information rather than acting simply as a regularizer, we also run an experiment with the \concat\ model using a random sentence as context instead of the previous sentence. This model achieves a BLEU of 31.65 and a pronoun F1 of 69.65 - both lower than the baseline, confirming that the extended context from the previous sentence does provide helpful information. 


\subsection{Fine-tuning on Pronoun-Targeted Data}
\label{subsection:data_finetuning}

{For the first set of fine-tuning experiments, we only fine-tune on the pronoun targeted subset $\gD_{\text{prn}}$ for the \sts\ model. This helps us assess the training schedule so that we can achieve a balance between preserving the information from the full data and gaining targeted information from the subset.} 


\vspace{-0.5em}
\paragraph{Setup.} Given a trained baseline model, we train additional epochs on the targeted subset $\gD_{\text{prn}}$. Apart from training only on the subset data, {we also try training on a shuffled dataset consisting of the training + targeted subset data (which essentially doubles the error-prone subset compared to the baseline training data)}, alternating the training between the subset and the full data ($\gD + \gD_{\text{prn}}$), and the subset and full data upsampled by 2 (\ie\ $2 \gD  + \gD_{\text{prn}}$).

To ensure that the results we see are from the fine-tuning and not simply from increased training, we train the original baseline model on the full data $\gD$ for additional epochs, equivalent to the number of fine-tuning epochs.

\vspace{-0.5em}
\paragraph{Results.} We see from the results in Table~\ref{table:additional_data_results} that although the pronoun F1 improves, the BLEU scores drop when the model is fine-tuned only with the subset data 
$\gD_{\text{prn}}$. {Shuffling a mix of the full training data with the subset data leads to a smaller drop in BLEU and a gain in pronoun F1}. However, alternating the training on the full corpus and the subset ($\gD  + \gD_{\text{prn}}$) stabilizes the BLEU score, and upsampling the primary dataset ($2 \gD +\gD_{\text{prn}}$) results in a smaller drop in BLEU, while {gaining more significantly} in pronoun F1 over the baseline. {A similar trend is also observed for the \concat\ model.} Further upsampling does not lead to a significant improvement in results, so all subsequent experiments upsample the primary dataset by 2.

{Increased training of the baseline also results in a drop in BLEU scores. However, the pronoun F1 is also lower, which is not the case for the fine-tuning results, indicating that fine-tuning rather than increased training is the source of the improvements.}


\begin{table}[t!]
\centering
\scalebox{0.70}{\begin{tabular}{lccccc}
\toprule
\textbf{Fine-tuning} & \textbf{WMT14}  & \multicolumn{4}{c}{\textbf{Pronoun Testset}} \\
 \cmidrule(lr){2-2} \cmidrule(lr){3-6} 
\textbf{data for \sts\ } & \textbf{BLEU} & \textbf{BLEU}  &\textbf{P} & \textbf{R} &\textbf{F1} \\

\midrule
  $\gD$ (baseline) & 31.64 & 35.56 & 77.92 & 66.01 & 69.55 \\
  \midrule
 $\gD_{\text{prn}}$ & 30.43 & 34.72 & 79.49 & 67.55 & 71.02 \\
  $\gD$  + $\gD_{\text{prn}}$ (shuffled) & 31.31 & 35.48 & 78.35 & 67.02 & 70.35\\
 $\gD$  + $\gD_{\text{prn}}$ & 31.23 & 35.39 & 79.61 & 67.99 & 71.40\\

 $2\gD $ + $\gD_{\text{prn}}$  & 31.56 & 35.57 & 79.25 & 68.01 & 71.35 \\
 $\gD$ (Increased training) & 31.53 & 35.60 & 78.14 & 66.15 & 69.77 \\

\midrule
  \multicolumn{5}{l}{\concat} \\
 \midrule
  $\gD$ (baseline) & 31.81 & 36.16 & 80.39 & 68.49 & 72.03 \\
  $2\gD $ + $\gD_{\text{prn}}$ & 31.31 & 36.12  & 81.20 & 69.35 & 72.84 \\
 
\bottomrule
\end{tabular}}
\vspace{-0.5em}
\caption{Subset data: fine-tuning results on the WMT14 De-En with precision, recall and F1 scores on the pronoun testset. $\gD$  represents the full training corpus; $2\gD$ is the full training corpus upsampled by 2, while $\gD_{\text{prn}}$ represents the pronoun targeted subset. }
\label{table:additional_data_results}

\end{table}

\subsection{Effect of Additional Losses}
\label{subsection:loss_finetuning}

{We conduct experiments using both targeted data and proposed hybrid losses.} 


\vspace{-0.2em}
\paragraph{Setup.} {In accordance with our settings to alternate training between the upsampled full dataset  and the subset data ($2\gD + \gD_{\text{prn}}$), we also alternate the additional loss such that it is only applied to the targeted subset. That is, in every alternate epoch, the model is trained on the upsampled full dataset $(2\gD)$ with the standard CLM translation loss $\gL_g$ (Eq.~\ref{eq:genloss}), and then trained on the targeted subset $\gD_{\text{prn}}$ with the proposed hybrid loss $\gL_{gd}$ (Eq.~\ref{eq:hybrid_obj}).}

{Each fine-tuning model is trained for 9 additional epochs, such that the first and the last epoch use the targeted subset data and loss. This is effectively about 4 cycles of fine-tuning on ($2\gD +\gD_{\text{prn}}$); further training does not lead to improved loss.}

\begin{table*}[t]

\begin{subtable}[t]{0.5\textwidth}
    \centering
     \scalebox{0.6}{\begin{tabular}{lc|ccccc}
\toprule
 & \textbf{Fine-tuning} & \textbf{WMT14}  & \multicolumn{4}{c}{\textbf{Pronoun Testset}} \\
 \cmidrule(lr){3-3} \cmidrule(lr){4-7} 
\textbf{Model} & \textbf{data}  & \textbf{BLEU} & \textbf{BLEU}  &\textbf{P} & \textbf{R} &\textbf{F1} \\ 
\midrule

Baseline \sts & - & 31.64 & 35.56 & 77.92 & 66.01 & 69.55  \\
Baseline \concat & - &  {31.81} & {36.16} & {80.39} & {68.49} & {72.03} \\

\midrule
\multicolumn{6}{c}{\textbf{All tokens}} \\
\midrule
 \sts &  $2\gD + \gD_{\text{prn}}$ & \textbf{32.14*} & 36.16 & 78.83 & 66.15 & 69.77* \\
 \sts & $2\gD + \gD_{\text{rand}}$ & 31.86 & 35.88 & 78.07 & 66.00 & 69.65 \\
 \sts & $\gD$ & 31.75 & 36.34 & 78.27 & 66.36 & 69.91 \\
 \concat & $2\gD + \gD_{\text{prn}}$ & 31.75 & \textbf{36.70} & \textbf{81.25} & \textbf{69.27} & \textbf{72.88} \\

 \midrule
 
 \multicolumn{6}{c}{\textbf{Only Pronouns}} \\
 \midrule

 \sts & $2\gD + \gD_{\text{prn}}$ & 31.81* & 36.43 & 78.62 & 66.82 & 70.37* \\
 \sts & $2\gD + \gD_{\text{rand}}$ & 31.71 & 36.12 & 78.65 & 66.72 & 70.32 \\
 \sts & $\gD$ & 31.89 & 36.20 &  78.31 & 66.32 & 69.98 \\
 \concat & $2\gD + \gD_{\text{prn}}$&  \textbf{31.99*} & \textbf{36.64} & \textbf{80.87} & \textbf{69.07} & \textbf{72.64} \\

\bottomrule

\end{tabular}

}
\caption{Fine-tuning results using \textbf{max-margin} loss.}
\label{subtable:pairwise_loss_results}
\end{subtable}
\hfill
\begin{subtable}[t]{0.5\textwidth}
\centering
     \scalebox{0.6}{\begin{tabular}{lc|ccccc}
\toprule
 & \textbf{Fine-tuning} & \textbf{WMT14}  & \multicolumn{4}{c}{\textbf{Pronoun Testset}} \\
 \cmidrule(lr){3-3} \cmidrule(lr){4-7} 
\textbf{Model} & \textbf{data}  & \textbf{BLEU} & \textbf{BLEU}  &\textbf{P} & \textbf{R} &\textbf{F1} \\ 
\midrule

Baseline \sts & - & 31.64 & 35.56 & 77.92 & 66.01 & 69.55  \\
Baseline \concat & - & {31.81} & {36.16} & {80.39} & {68.49} & {72.03} \\

\midrule
\multicolumn{6}{c}{\textbf{All tokens}} \\
\midrule 
 \sts & $2\gD + \gD_{\text{prn}}$ & 31.83* & 36.50 & 79.18 & 67.16 & 70.78* \\
 \sts & $2\gD + \gD_{\text{rand}}$ & 31.73 & 36.16 & 78.32 & 66.62 & 70.15 \\
 \sts & $\gD$ & 31.77  & 36.24 &  78.35 & 66.17 & 69.86 \\
 \concat & $2\gD + \gD_{\text{prn}}$ & \textbf{31.85} & \textbf{36.61}  & \textbf{80.91} & \textbf{68.91} & \textbf{72.57} \\

 \midrule
 
 \multicolumn{6}{c}{\textbf{Only Pronouns}} \\
 \midrule

 \sts & $2\gD + \gD_{\text{prn}}$ &  31.73 & 36.30 & 79.01 & 66.80 & 70.50* \\
 \sts  & $2\gD + \gD_{\text{rand}}$ & \textbf{32.05} & 36.43  & 78.35 & 66.25 & 69.87 \\
 \sts & $\gD$ & \textbf{32.05} & 35.81 & 78.58 & 66.52 & 70.22 \\
 \concat & $2\gD + \gD_{\text{prn}}$ & \textbf{32.00*} & \textbf{36.57} & \textbf{80.89} & \textbf{68.66} & \textbf{72.39} \\
\bottomrule
\end{tabular}}

\caption{Fine-tuning results using \textbf{log-likehihood} loss}
\label{subtable:nll_loss_results}
\end{subtable}
\vspace{-0.5em}
 \caption{{Targeted fine-tuning loss: fine-tuning results on the WMT14 De-En testset with F1 scores on the pronoun testset. Fine-tuning results on $2\gD + \gD_{\text{prn}}$ refer to alternated training with pronoun-targeted fine-tuning data and the upsampled full training data. Fine-tuning on $2\gD + \gD_{\text{rand}}$ is the same setting with the targeted data replaced with a random subset. Fine-tuning on $\gD$ refers to additional training with the hybrid losses applied on the full dataset. * indicates statistically significant difference from the baseline (p $\leq$ 0.05 for F1; $>$80\% confidence for BLEU). }}
\end{table*}

{Apart from applying the additional loss on all tokens in the subset data, we also experiment with applying the additional loss only on the pronoun tokens, \ie\ the loss is only applied to those tokens which have a pronoun as the target translation.}

{To further assess the contribution of the targeted subset data, we conduct experiments by replacing it with a random subset $\gD_{\text{rand}}$. We also conduct fine-tuning experiments by applying the additional loss on the full training dataset $\gD$ while training the baseline model for additional epochs.}


 





\vspace{-0.5em}
\paragraph{Max-margin loss results.} {Results for fine-tuning with the max-margin loss are shown in Table~\ref{subtable:pairwise_loss_results}. We see that there is an improvement in BLEU from 31.64 to \textbf{32.14}. From the difference in improvement of the results from fine-tuning over $\gD_{\text{rand}}$ and $\gD$, it is apparent that this increase is a consequence of both the targeted data and the targeted loss. There is also a corresponding increase in pronoun F1 from 69.55 to 69.77.}

{More importantly, we see that the \concat\ model drops slightly in BLEU to 31.75 with respect to the baseline, but the pronoun translation F1 improves from 72.03 to \textbf{72.88}. When the loss is applied only on pronouns, the \sts\ model has a smaller BLEU increase to 31.81, but a larger pronoun F1 increase to 70.37. The \concat\ model benefits the most from having both pronoun-targeted fine-tuning data and loss; both the BLEU score and the pronoun F1 improve. }


 





\vspace{-0.5em}
\paragraph{Log-likelihood loss results.} {Results for fine-tuning with the log-likelihood loss are shown in Table~\ref{subtable:nll_loss_results}. The overall increase in BLEU with the log-likelihood loss is lower for \sts\  compared to the max-margin loss, but the improvements in pronoun F1 are higher. With respect to the results on fine-tuning over $\gD_{\text{rand}}$ and $\gD$, improvement in BLEU score here does not result in a corresponding improvement in pronoun translation, further confirming the contribution of the targeted data. Once again, the \concat\ model outperforms the rest by gaining in both BLEU and pronoun F1.}

{Both losses perform comparably - while the \sts\ model achieves a higher increase in BLEU with the max-margin loss, gains in pronoun translation are higher with the log-likelihood loss. For the \concat\ model, both losses provide similar BLEU improvements, but the max-margin loss leads to higher gains in pronoun F1. }





 





\section{Additional Experiments and Analysis}
\label{sec:analysis}
\subsection{Qualitative Analysis of Results}

\begin{table*}[t]
\small 
\centering
\scalebox{0.75}{\begin{tabular}{ll}
\toprule
\textbf{Description} & \multicolumn{1}{c}{\textbf{Examples}} \\
\midrule
\multicolumn{2}{c}{\textbf{WMT14 Testset}} \\
\midrule

Source & 14 stunden kämpften die ärzte um das überleben des opfers , jedoch vergeblich .\\
Reference & for 14 hours, doctors battled to save the life of the victim , ultimately in vain .\\
Baseline &  \red{14 hours of doctors} fought for the victim's survival , but in vain .
\\
Our best model & \blue{the doctors fought 14 hours} for the survival of the victim , but in vain . \\
 \midrule
 
Source & der handel am nasdaq options market wurde am freitagnachmittag deutscher zeit unterbrochen . \\

Reference & trading at the nasdaq options market was interrupted on friday afternoon , german time .\\
 
Baseline & trade at nasdaq options market was cut off \red{on the german friday afternoon .}\\

Our best model & trade in nasdaq options market was suspended \blue{on friday afternoon in germany .}\\
 




\midrule
\multicolumn{2}{c}{\textbf{Pronoun Testset}} \\
\midrule

Context & ... die die amerikanische flamme in die umnachtete welt bringe : lady liberty geht voran .\\
Source & sie soll die fackel der freiheit von den vereinigten staaten in den rest der welt tragen . \\
Context & ... taking the american flame out to the benighted world : \textbf{lady liberty} is stepping forward . \\
Reference & she is meant to be carrying the torch of liberty from the united states to the rest of the world .\\
Baseline & \red{it} is meant to carry the torch of freedom from the united states to the rest of the world . \\
Our best model & \blue{she} is supposed to carry the torch of freedom from the united states to the rest of the world . \\

\midrule


Context & versteinerte reste der haut bedecken noch immer die holprigen panzerplatten , die den schädel des tieres tragen . \\
Source &  sein rechter vorderfuß liegt an seiner seite , seine fünf finger sind nach oben gespreizt . \\
Context & fossilized remnants of skin still cover the bumpy armor plates dotting the \textbf{animal's} skull .\\
Reference &  its right forefoot lies by its side , its five digits splayed upward . \\
Baseline & \red{his} right - hand front foot is on \red{his} side , \red{his} five fingers are spiked up .\\
Our best model & \blue{its} right front foot is on \blue{its} side , \blue{its} five fingers are split upwards . \\

\bottomrule
\end{tabular}}
\vspace{-0.5em}
\caption{Examples showing the improvements in translations from our best models, across the WMT14 and the pronoun testsets. The previous sentence context information for the pronoun testset is also shown. }
\label{table:examples_1}

\end{table*}

We performed a qualitative analysis to see the effect of our fine-tuning. Some examples of improvements in translation resulting from our fine-tuning are shown in Table~\ref{table:examples_1} (see Appendix for more).

The results of the targeted fine-tuning show that both the targeted data and the additional loss play a role in improving the translations. Another important conclusion that can be drawn is that there is no correlation between the BLEU score and the pronoun translation quality; in this case we have shown that it is possible to target the improvement of pronoun translations. 

{However, for the \sts\ model in particular, we see that there are improvements in BLEU that do not correspondingly improve pronoun translations, which can be surprising given that the fine-tuning data is targeted towards pronouns. It can be surmised from the improvements in the \concat\ model that the \sts\ model fails to improve in pronoun translation because it simply lacks the additional information that the context provides, which can be important for translating discourse phenomena like pronouns \cite{SennrichDisourceMT}. See Table~\ref{table:examples_1} for examples from the pronoun testset.}

{Another anomaly is that in some cases, the pronoun translation results are better when the loss is applied to all tokens rather than only to pronouns. A similar phenomenon may be the cause here - improved translation of the rest of the sentence may result in better contextual information, that in turn leads to better pronoun translations. This underscores the importance of using context rather than trying to improve pronoun translations in isolation.  }

The general improvements in BLEU result from the fact that the targeted data is a subset that the model has failed to learn adequately from. Thus, our method of obtaining targeted data seemingly results in a subset that is generally poorly translated by the original baseline model, so training on this data results in an improved BLEU score. {This also explains the disparity in results with the fine-tuning on the random ($\gD_{\text{rand}}$) and the full ($\gD$) datasets.} 

\subsection{Comparison with Related Work}

\begin{table}[t]
\centering
\scalebox{0.65}{\begin{tabular}{lccccc}
\toprule
 & \textbf{WMT14}  & \multicolumn{4}{c}{\textbf{Pronoun Testset}} \\
  \cmidrule(lr){2-2} \cmidrule(lr){3-6} 
\textbf{Model} & \textbf{BLEU} & \textbf{BLEU}  &\textbf{P} & \textbf{R} &\textbf{F1} \\ 
\midrule
Baseline \sts & 31.64 & 35.56 & 77.92 & 66.01 & 69.55 \\
DocRepair$^*$ & 30.07 & 32.58 & 77.29 & 64.46 & 68.36 \\
Backtranslation$^*$ & \textbf{32.57} & \textbf{38.54} & 80.61 & 67.14 & 71.37 \\
Best fine-tuned \sts\ & 32.14 & 36.16 & 78.83 & 66.15 & 69.77 \\
Best fine-tuned \concat & {32.00} & {36.57} & \textbf{80.89} & \textbf{68.66} & \textbf{72.39} \\
\bottomrule
\end{tabular}}
\caption{Comparison with backtranslation and the DocRepair post-editing model. $^*$ indicates models use extra monolingual data. BLEU scores reported on the WMT14 De-En testset, with \textbf{P}recision/\textbf{R}ecall/\textbf{F1} on the pronoun testset. For DocRepair, the input is the output from our baseline \sts\ De-En model.}
\label{table:comparison_results}

\end{table}

\paragraph{Backtranslation.}

We train a target-source En-De model with the same training data ($\gD$, consisting of 2.5M pairs of parallel data) and settings as the baseline \sts\ model. This achieves a BLEU score of \textbf{27.4} on the WMT14 En-De testset. We use this model to translate about 76M sentences of NewsCrawl, a monolingual English corpus, to German. Using this pseudo-parallel corpus in addition to the original training corpus ($\approx$ 78M pairs), we train a \sts\ source-target De-En backtranslation model. This model is trained for 500K steps. The results are shown in Table~\ref{table:comparison_results}. Although backtranslation achieves highest BLEU score at 32.57, our fine-tuned \concat\ model achieves the highest F1 for pronoun translation at 72.39, {without having been trained on any extra monolingual data.} This is further proof that it may be insufficient to simply improve the BLEU scores at a sentence-level. {Performing fine-tuning on a \concat\ backtranslation model may be interesting to consider; we leave this for future work.}\footnote{{A caveat here is that this would require training alternately on a targeted subset and an upsampled backtranslation dataset according to our training schedule. Considering the size of the backtranslation dataset, it would require massive amounts of additional training.}}

%

\paragraph{Automatic post-editing.} {We train a contextual, monolingual automatic post-editing model proposed by \citet{Voita2019ContextAwareMR} for English. To capture MT errors, the model is trained with round-trip-translated texts as inputs with reference texts as the intended outputs. We use default settings and {similar} data sizes as proposed in their paper. We use 2.5M sentences from parallel data $\gD$ and monolingual English sentences from NewsCrawl to make up $\approx$ 30M sentences. Using the En-De model described above and our baseline De-En model, we translate this data  to German and then back to English to obtain round-trip translations. We use this data to train their model\footnote{Taken from \href{https://github.com/lena-voita/good-translation-wrong-in-context}{https://github.com/lena-voita/good-translation-wrong-in-context}.} for around 750K steps as recommended by the authors.}

{We use the outputs of our baseline \sts\ De-En model on the WMT14 testset and the pronoun challenge testset as input to the model.\footnote{For the pronoun testset, we were only able to provide groups of 3 sentences as input instead of 4 which the original model uses, since the testset only provides two previous sentences as context. {We add dummy text as the first sentence to make it a 4-sentence group input.}} The results are shown in Table~\ref{table:comparison_results}. We see that automatic post-editing does not lead to an improvement in BLEU\footnote{Note that we calculate the BLEU scores for each sentence separately as is standard, unlike in groups of 4 as the original paper. This is to more accurately compare against the results from the rest of our experiments.} or pronoun translation in this case.} 

Our analysis of round-trip-translations suggests that this is possibly because they do not contain enough errors.
Experiments conducted on the WMT14 En-De testset show that
if it is translated using our En-De model (BLEU:27.40) to German and then translated using our De-En model (BLEU:31.64) back to English, the resulting text has a BLEU of 44.44, which is significantly higher. It is a well-known phenomenon that MT models perform substantially better on \textit{translationese} \cite{Graham2019TranslationeseIM}, which refers to text that is unnatural by virtue of being translated. This means that it is not very likely to resemble typical MT output or capture the same errors \cite{Poncelas2018InvestigatingBI}; twice-translated texts therefore contain considerably fewer errors that can be learnt from. 



\subsection{Results on the IWSLT13 Testset}
\label{subsec:iwslt}
We evaluate our fine-tuned models on the IWSLT13 De-En testset (Table~\ref{table:others_BLEU_results}).  We also evaluate the {pronoun translation} for this testset. The backtranslation model fails to generalize, and performs worse than the baseline. It can be seen that our fine-tuned models improve over the baseline performance on this testset as well; the best \sts\ model improves from 31.64 to 32.16, while the best \concat\ model improves from 32.10 to 33.13, with corresponding improvements in pronoun F1. \concat\ continues to be the best performing model, showing significant improvements for both fine-tuning losses.

\begin{table}[t]
\centering
\scalebox{0.75}{\begin{tabular}{ccccc}
\toprule
 &\multicolumn{2}{c}{\sts}  & \multicolumn{2}{c}{\concat}  \\
  \cmidrule(lr){2-3} \cmidrule(lr){4-5} 
 \textbf{Model} & \textbf{BLEU} & \textbf{Prn. F1} & \textbf{BLEU} & \textbf{Prn. F1} \\
\midrule
Baseline & 31.64 & 60.47 & 32.10 & 62.01 \\
Backtranslation & 30.30 & 58.02 & - & -\\
\midrule
\multicolumn{5}{c}{\textbf{All tokens}} \\
\midrule 
Max-margin &  31.88 & 60.87 & \textbf{32.95} & 61.90 \\
Log-likelihood & 32.02 & 60.64 & 32.78  & \textbf{62.10}\\

 \midrule
 
 \multicolumn{5}{c}{\textbf{Only Pronouns}} \\
 \midrule
Max-margin & 32.13 & 60.61 & \textbf{33.13} & \textbf{62.20} \\
Log-likelihood & 32.16 & 60.83 & 32.78 & 61.97 \\

\bottomrule
\end{tabular}}
\caption{BLEU score and \textbf{Pr}o\textbf{n}oun translation \textbf{F1} results of the baselines and the fine-tuned models on the IWSLT13 De-En testset. }
\label{table:others_BLEU_results}

\end{table}

\subsection{Generalizability to Other Languages}
\label{subsec:other_langs}
{Finally, we test the generalizability of our fine-tuning method by running experiments for French-English and Czech-English. We use the same training dataset sources as for German-English (\ie\ News Commentary, IWSLT \cite{IWSLT} and Europarl \cite{Europarl}). This results in 2.53M sentences of training data and 500K sentences of fine-tuning data for Fr-En, and 992K sentences of training data and 100K sentences of fine-tuning data for Cs-En. We report the baseline BLEU results on the WMT14 testsets and the pronoun translation results on the corresponding testsets from \citet{EvalPronom} containing 1478 (Fr-En) and 1686 (Cs-En) sentences. We see from Table~\ref{tab:add_source_langs} that our fine-tuning approach shows similar trends in improving BLEU and pronoun translation results for both Fr-En and Cs-En.  }

\begin{table*}[t]

\begin{subtable}[t]{0.5\textwidth}
    \centering
     \scalebox{0.6}{\begin{tabular}{lc|ccccc}
\toprule
 & \textbf{Fine-tuning} & \textbf{WMT14}  & \multicolumn{4}{c}{\textbf{Pronoun Testset}} \\
 \cmidrule(lr){3-3} \cmidrule(lr){4-7} 
\textbf{Model} & \textbf{loss}  & \textbf{BLEU} & \textbf{BLEU}  &\textbf{P} & \textbf{R} &\textbf{F1} \\ 
\midrule
Baseline \sts & - & 35.61 & 34.53 & 90.64 & 64.00 & 73.73  \\
Baseline \concat & - &  {36.06} & {35.18} & {84.86} & {72.07} & {75.86} \\

\midrule
\multicolumn{6}{c}{\textbf{All tokens}} \\
\midrule
\sts\ & max-margin & \textbf{36.12}* & 35.31 & 93.61 & 64.26 & 74.56* \\
\sts\ & log-likelihood & 36.04* & 35.39 & 96.39 & 66.95 & \textbf{77.38}* \\
\concat\ & max-margin & 35.98 & 35.41 & 85.93 & 72.48 & 76.48 \\
\concat & log-likelihood & 35.98 & 35.09 & 85.07 & 71.43 & 75.51 \\

 \midrule
 
 \multicolumn{6}{c}{\textbf{Only Pronouns}} \\
 \midrule
\sts\ & max-margin & 36.05* & 35.34 & 93.48 & 67.24 & 76.96 \\
\sts\ & log-likelihood & 35.86* & 35.09 & 93.62 & 63.74 & 73.88 \\
\concat & max-margin & 35.97 & 35.26 & 85.71 & 71.97 & 76.07 \\
\concat & log-likelihood & \textbf{36.09} & 35.55 & 85.85 & 72.38 & \textbf{76.50} \\

\bottomrule

\end{tabular}

}
\caption{Fine-tuning results for French-English}
\label{subtable:pairwise_loss_results}
\end{subtable}
\hfill
\begin{subtable}[t]{0.5\textwidth}
\centering
     \scalebox{0.6}{\begin{tabular}{lc|ccccc}
     
\toprule
& \textbf{Fine-tuning} & \textbf{WMT14}  & \multicolumn{4}{c}{\textbf{Pronoun Testset}} \\
 \cmidrule(lr){3-3} \cmidrule(lr){4-7} 
\textbf{Model} & \textbf{loss}  & \textbf{BLEU} & \textbf{BLEU}  &\textbf{P} & \textbf{R} &\textbf{F1} \\
\midrule

Baseline \sts\ & - & 25.23 & 21.88 & 82.65 & 48.78 & 60.40 \\
Baseline \concat\ & - & 28.27 & 24.19 & 71.94 & 55.57 & 60.37 \\

\midrule
\multicolumn{6}{c}{\textbf{All tokens}} \\
\midrule 
\sts\ & max-margin & \textbf{26.13}* & 22.49 & 84.18 & 50.71 & \textbf{62.16}* \\
\sts\ & log-likehood & 26.08* & 22.65 & 83.02 & 49.02 & 60.53 \\
\concat & max-margin & 27.56 & 23.69 & 73.82 & 57.81 & 62.45* \\
\concat & log-likelihood & 27.50 & 23.85 & 74.43 & 58.17 & \textbf{62.89}* \\

 \midrule
 
 \multicolumn{6}{c}{\textbf{Only Pronouns}} \\
 \midrule
\sts & max-margin & 26.10* & 22.56 & 83.02 & 49.96 & 61.03 \\
\sts & log-likelihood & 26.01* & 22.62 & 83.90 & 49.17 & 60.88 \\
\concat & max-margin & 27.48 & 23.76 & 74.20 & 57.72 & 62.53* \\
\concat & log-likelihood & 27.59 & 23.72 & 74.18 & 57.77 & 62.54 \\

\bottomrule
\end{tabular}}

\caption{Fine-tuning results for Czech-English}
\label{subtable:nll_loss_results}
\end{subtable}
\caption{Results for experiments on generalizability to other source languages, Fr-En and Cs-En. * indicates results are statistically significant.}
\label{tab:add_source_langs}
\end{table*}

\subsection{Discussion}
{Our objective is to propose a novel fine-tuning method that leverages “unlearned” data using additional loss. To this end, we proposed two different losses. We do not mean to advocate for any particular loss; in our experiments we happened to get comparable results, which may not conclusively point to one loss as being better. A different loss may perform better in other tasks.}

{Although we focused on pronoun translations, our fine-tuning method is generic and can be used to correct other kinds of errors in machine translations, like named entities or other rare words. Our proposed losses can be adapted to other directed generation tasks; \eg\, to improve coherence/factual correctness in abstractive summarization, or for controlled text generation. Our fine-tuning approach also opens up new ways to address training issues that originate from datasets; \eg\, it could potentially be used to correct biases (such as gender) or used to improve system robustness.}

\section{Related Work}
\label{sec:relwork}
    

Our idea of conditional generative-discriminative training is related to the idea of discriminative training of generative models. Previously, this idea was proposed for Markov models. \citet{collins-2002-discriminative} trained a Hidden Markov Model (HMM) discriminatively for sequence tagging with structured perceptron algorithm. \citet{yakhnenko2005discriminatively} used a similar idea for sequence classification. In deep learning, the well-known generative adversarial networks (GANs) \cite{goodfellow2014generative} are an example where a generator is trained with the help of a discriminator. To the best of our knowledge, ours is the first work to explore this idea with conditional language models for guiding the model on what to generate and what not to generate.  

A few fine-tuning methods are related to our work. \citet{abdulmumin2019tagless} pre-train an MT model on synthetic backtranslated data and fine-tune it on authentic parallel data, and show that it can improve 0.7 BLEU over backtranslation on English-Vietnamese. \citet{Fadaee2018BackTranslationSB} use various sampling strategies to improve the results of backtranslation by targeting difficult-to-predict words based on prediction loss. Our strategy is similar in that we also try to target words that the model has trouble with, but we do not use additional data.

A number of methods have been proposed for adapting a trained MT model to another domain by fine-tuning. A common strategy is to simply perform additional training on the new domain dataset \cite{LuongManningiwslt15} or use a mix of in-domain and out-domain data for fine-tuning without loss of generalization \cite{Chu2017AnEC} or upweight out-of-domain data \cite{Wang2017InstanceWF}. 

There has been some work on targeted improvement of translations, specifically for  named-entities. \citet{ugawa-etal-2018-neural} adapt MT network architecture to encode named entity features and tags while \citet{li-etal-2018-named} perform domain adaptation in addition to feature encoding. {With respect to discourse phenomena, \citet{Stojanovski2019ImprovingAR} propose a curriculum learning based approach, where a context-aware model is trained on randomly sampled oracle data containing gold-standard pronouns. In our work, we focus on the baseline model's failings and try to increase its learning capacity by proposing additional losses. }

Most recent work on improving pronoun translations has involved building more complex architectures that incorporate contextual information \cite{Voita-ACL18, Wong2020ContextualNM}. In contrast, we present a more generalized approach.





\section{Conclusions and Future Work}
\label{sec:conc}
We have proposed a class of conditional generative-discriminative losses to increase the learning potential of NMT models, showing that it is possible to leverage ``unlearned" training data to further improve an MT model, by strategically filtering the data and applying additional targeted losses. 

{We demonstrated the effectiveness of our methods on {different languages and testsets,} also reporting improved pronoun translations. Although we focus on pronoun translations, our fine-tuning method is generic and can be used to correct other kinds of errors in machine translations, like named entities or other rare words. In future work, we will explore other such applications of our proposed methods.  }



\section*{Acknowledgments}

We would like to thank the anonymous reviewers for their helpful comments. Shafiq Joty would like to thank the funding support from NRF (NRF2016IDM-TRANS001-062), Singapore.


\bibliographystyle{acl_natbib}
\bibliography{anthology,finetune}

\begin{thebibliography}{44}
\expandafter\ifx\csname natexlab\endcsname\relax\def\natexlab#1{#1}\fi

\bibitem[{Abdulmumin et~al.(2019)Abdulmumin, Galadanci, and
  Garba}]{abdulmumin2019tagless}
Idris Abdulmumin, B.~S. Galadanci, and Aliyu~Dadan Garba. 2019.
\newblock Tag-less back-translation.
\newblock \emph{ArXiv}, abs/1912.10514.

\bibitem[{Bahdanau et~al.(2015)Bahdanau, Cho, and
  Bengio}]{Bahdanau2015NeuralMT}
Dzmitry Bahdanau, Kyunghyun Cho, and Yoshua Bengio. 2015.
\newblock Neural machine translation by jointly learning to align and
  translate.
\newblock \emph{ArXiv}, abs/1409.0473.

\bibitem[{Caswell et~al.(2019)Caswell, Chelba, and
  Grangier}]{Caswell2019TaggedB}
Isaac Caswell, Ciprian Chelba, and David Grangier. 2019.
\newblock \href {https://doi.org/10.18653/v1/W19-5206} {Tagged
  back-translation}.
\newblock In \emph{Proceedings of the Fourth Conference on Machine Translation
  (Volume 1: Research Papers)}, pages 53--63, Florence, Italy. Association for
  Computational Linguistics.

\bibitem[{Cettolo et~al.(2012)Cettolo, Girardi, and Federico}]{IWSLT}
Mauro Cettolo, Christian Girardi, and Marcello Federico. 2012.
\newblock Wit$^3$: Web inventory of transcribed and translated talks.
\newblock In \emph{Proceedings of the 16$^{th}$ Conference of the European
  Association for Machine Translation (EAMT)}, pages 261--268, Trento, Italy.

\bibitem[{Chu et~al.(2017)Chu, Dabre, and Kurohashi}]{Chu2017AnEC}
Chenhui Chu, Raj Dabre, and Sadao Kurohashi. 2017.
\newblock An empirical comparison of simple domain adaptation methods for
  neural machine translation.
\newblock \emph{ArXiv}, abs/1701.03214.

\bibitem[{Collins(2002)}]{collins-2002-discriminative}
Michael Collins. 2002.
\newblock \href {https://doi.org/10.3115/1118693.1118694} {Discriminative
  training methods for hidden {M}arkov models: Theory and experiments with
  perceptron algorithms}.
\newblock In \emph{Proceedings of the 2002 Conference on Empirical Methods in
  Natural Language Processing ({EMNLP} 2002)}, pages 1--8. Association for
  Computational Linguistics.

\bibitem[{Collobert et~al.(2011)Collobert, Weston, Bottou, Karlen, Kavukcuoglu,
  and Kuksa}]{collobert2011natural}
Ronan Collobert, Jason Weston, L{\'e}on Bottou, Michael Karlen, Koray
  Kavukcuoglu, and Pavel Kuksa. 2011.
\newblock Natural language processing (almost) from scratch.
\newblock \emph{The Journal of Machine Learning Research}, 12:2493--2537.

\bibitem[{Edunov et~al.(2018)Edunov, Ott, Auli, and
  Grangier}]{EdunovBacktranslationatscale}
Sergey Edunov, Myle Ott, Michael Auli, and David Grangier. 2018.
\newblock \href {https://doi.org/10.18653/v1/D18-1045} {Understanding
  back-translation at scale}.
\newblock In \emph{Proceedings of the 2018 Conference on Empirical Methods in
  Natural Language Processing}, pages 489--500, Brussels, Belgium. Association
  for Computational Linguistics.

\bibitem[{Edunov et~al.(2020)Edunov, Ott, Ranzato, and
  Auli}]{eval_back_translation_translationese}
Sergey Edunov, Myle Ott, Marc{'}Aurelio Ranzato, and Michael Auli. 2020.
\newblock \href {https://doi.org/10.18653/v1/2020.acl-main.253} {On the
  evaluation of machine translation systems trained with back-translation}.
\newblock In \emph{Proceedings of the 58th Annual Meeting of the Association
  for Computational Linguistics}, pages 2836--2846, Online. Association for
  Computational Linguistics.

\bibitem[{Fadaee and Monz(2018)}]{Fadaee2018BackTranslationSB}
Marzieh Fadaee and Christof Monz. 2018.
\newblock \href {https://doi.org/10.18653/v1/D18-1040} {Back-translation
  sampling by targeting difficult words in neural machine translation}.
\newblock In \emph{Proceedings of the 2018 Conference on Empirical Methods in
  Natural Language Processing}, pages 436--446, Brussels, Belgium. Association
  for Computational Linguistics.

\bibitem[{Freitag et~al.(2019)Freitag, Caswell, and Roy}]{Freitag2019TextRM}
Markus Freitag, Isaac Caswell, and Scott Roy. 2019.
\newblock Text repair model for neural machine translation.
\newblock \emph{ArXiv}, abs/1904.04790.

\bibitem[{Goodfellow et~al.(2014)Goodfellow, Pouget-Abadie, Mirza, Xu,
  Warde-Farley, Ozair, Courville, and Bengio}]{goodfellow2014generative}
Ian Goodfellow, Jean Pouget-Abadie, Mehdi Mirza, Bing Xu, David Warde-Farley,
  Sherjil Ozair, Aaron Courville, and Yoshua Bengio. 2014.
\newblock Generative adversarial nets.
\newblock In \emph{Advances in neural information processing systems}, pages
  2672--2680.

\bibitem[{Graham et~al.(2019)Graham, Haddow, and
  Koehn}]{Graham2019TranslationeseIM}
Yvette Graham, Barry Haddow, and Philipp Koehn. 2019.
\newblock Translationese in machine translation evaluation.
\newblock \emph{ArXiv}, abs/1906.09833.

\bibitem[{Guillou(2012)}]{Guillou2012ImprovingPT}
Liane Guillou. 2012.
\newblock \href {https://www.aclweb.org/anthology/E12-3001} {Improving pronoun
  translation for statistical machine translation}.
\newblock In \emph{Proceedings of the Student Research Workshop at the 13th
  Conference of the {E}uropean Chapter of the Association for Computational
  Linguistics}, pages 1--10, Avignon, France. Association for Computational
  Linguistics.

\bibitem[{Çaglar G{\"u}lçehre et~al.(2017)Çaglar G{\"u}lçehre, Firat, Xu,
  Cho, and Bengio}]{Glehre2017LanguageModelNMT}
Çaglar G{\"u}lçehre, Orhan Firat, Kelvin Xu, Kyunghyun Cho, and Yoshua
  Bengio. 2017.
\newblock On integrating a language model into neural machine translation.
\newblock \emph{Comput. Speech Lang.}, 45:137--148.

\bibitem[{Hardmeier and Federico(2010)}]{HardmeierAutoPRF}
Christian Hardmeier and Marcello Federico. 2010.
\newblock Modelling pronominal anaphora in statistical machine translation.
\newblock In \emph{Proceedings of the 2010 International Workshop on Spoken
  Language Translation}, IWSLT~'10, pages 283--289, Paris, France.

\bibitem[{Hoang et~al.(2018)Hoang, Koehn, Haffari, and
  Cohn}]{Hoang2018IterativeBacktranslation}
Vu~Cong~Duy Hoang, Philipp Koehn, Gholamreza Haffari, and Trevor Cohn. 2018.
\newblock \href {https://doi.org/10.18653/v1/W18-2703} {Iterative
  back-translation for neural machine translation}.
\newblock In \emph{Proceedings of the 2nd Workshop on Neural Machine
  Translation and Generation}, pages 18--24, Melbourne, Australia. Association
  for Computational Linguistics.

\bibitem[{Jwalapuram et~al.(2019)Jwalapuram, Joty, Temnikova, and
  Nakov}]{EvalPronom}
Prathyusha Jwalapuram, Shafiq Joty, Irina Temnikova, and Preslav Nakov. 2019.
\newblock \href {https://doi.org/10.18653/v1/D19-1294} {Evaluating pronominal
  anaphora in machine translation: An evaluation measure and a test suite}.
\newblock In \emph{Proceedings of the 2019 Conference on Empirical Methods in
  Natural Language Processing and the 9th International Joint Conference on
  Natural Language Processing (EMNLP-IJCNLP)}, pages 2964--2975, Hong Kong,
  China. Association for Computational Linguistics.

\bibitem[{Jwalapuram et~al.(2020)Jwalapuram, Rychalska, Joty, and
  Basaj}]{Jwalapuram2020CanYC}
Prathyusha Jwalapuram, Barbara Rychalska, Shafiq~R. Joty, and Dominika Basaj.
  2020.
\newblock Can your context-aware {MT} system pass the {DiP} benchmark tests? :
  {E}valuation benchmarks for discourse phenomena in machine translation.
\newblock \emph{ArXiv}, abs/2004.14607.

\bibitem[{Koehn and Knowles(2017)}]{koehn-knowles-2017-six}
Philipp Koehn and Rebecca Knowles. 2017.
\newblock \href {https://doi.org/10.18653/v1/W17-3204} {Six challenges for
  neural machine translation}.
\newblock In \emph{Proceedings of the First Workshop on Neural Machine
  Translation}, pages 28--39, Vancouver. Association for Computational
  Linguistics.

\bibitem[{L{\"a}ubli et~al.(2018)L{\"a}ubli, Sennrich, and
  Volk}]{Lubli2018HasMT}
Samuel L{\"a}ubli, Rico Sennrich, and Martin Volk. 2018.
\newblock \href {https://doi.org/10.18653/v1/D18-1512} {Has machine translation
  achieved human parity? a case for document-level evaluation}.
\newblock In \emph{Proceedings of the 2018 Conference on Empirical Methods in
  Natural Language Processing}, pages 4791--4796, Brussels, Belgium.
  Association for Computational Linguistics.

\bibitem[{Lewis et~al.(2020)Lewis, Liu, Goyal, Ghazvininejad, Mohamed, Levy,
  Stoyanov, and Zettlemoyer}]{Lewis2019BARTDS}
Mike Lewis, Yinhan Liu, Naman Goyal, Marjan Ghazvininejad, Abdelrahman Mohamed,
  Omer Levy, Veselin Stoyanov, and Luke Zettlemoyer. 2020.
\newblock \href {https://doi.org/10.18653/v1/2020.acl-main.703} {{BART}:
  Denoising sequence-to-sequence pre-training for natural language generation,
  translation, and comprehension}.
\newblock In \emph{Proceedings of the 58th Annual Meeting of the Association
  for Computational Linguistics}, pages 7871--7880, Online. Association for
  Computational Linguistics.

\bibitem[{Li et~al.(2018)Li, Wang, Aw, Chng, and Li}]{li-etal-2018-named}
Zhongwei Li, Xuancong Wang, Ai~Ti Aw, Eng~Siong Chng, and Haizhou Li. 2018.
\newblock \href {https://doi.org/10.18653/v1/W18-2407} {Named-entity tagging
  and domain adaptation for better customized translation}.
\newblock In \emph{Proceedings of the Seventh Named Entities Workshop}, pages
  41--46, Melbourne, Australia. Association for Computational Linguistics.

\bibitem[{Luong and Manning(2015)}]{LuongManningiwslt15}
Minh-Thang Luong and Christopher~D. Manning. 2015.
\newblock Stanford neural machine translation systems for spoken language
  domain.
\newblock In \emph{International Workshop on Spoken Language Translation}, Da
  Nang, Vietnam.

\bibitem[{Miculicich et~al.(2018)Miculicich, Ram, Pappas, and
  Henderson}]{miculicich-etal-2018-document}
Lesly Miculicich, Dhananjay Ram, Nikolaos Pappas, and James Henderson. 2018.
\newblock \href {https://doi.org/10.18653/v1/D18-1325} {Document-level neural
  machine translation with hierarchical attention networks}.
\newblock In \emph{Proceedings of the 2018 Conference on Empirical Methods in
  Natural Language Processing}, pages 2947--2954, Brussels, Belgium.
  Association for Computational Linguistics.

\bibitem[{Nguyen et~al.(2020)Nguyen, Joty, Kui, and Aw}]{nguyen2019data}
Xuan-Phi Nguyen, Shafiq Joty, Wu~Kui, and Ai~Ti Aw. 2020.
\newblock Data diversification: A simple strategy for neural machine
  translation.
\newblock In \emph{Advances in Neural Information Processing Systems 34}.
  Curran Associates, Inc.

\bibitem[{{\"O}stling and Tiedemann(2016)}]{ostling2016efmaral}
Robert {\"O}stling and J{\"o}rg Tiedemann. 2016.
\newblock \href {http://ufal.mff.cuni.cz/pbml/106/art-ostling-tiedemann.pdf}
  {Efficient word alignment with {M}arkov {C}hain {M}onte {C}arlo}.
\newblock \emph{Prague Bulletin of Mathematical Linguistics}, 106:125--146.

\bibitem[{Poncelas et~al.(2018)Poncelas, Shterionov, Way, de~Buy~Wenniger, and
  Passban}]{Poncelas2018InvestigatingBI}
Alberto Poncelas, Dimitar~Sht. Shterionov, Andy Way, Gideon~Maillette
  de~Buy~Wenniger, and Peyman Passban. 2018.
\newblock Investigating backtranslation in neural machine translation.
\newblock \emph{ArXiv}, abs/1804.06189.

\bibitem[{Sennrich(2018)}]{SennrichDisourceMT}
Rico Sennrich. 2018.
\newblock \href {http://homepages.inf.ed.ac.uk/rsennric/wnmt2018.pdf} {Why the
  time is ripe for discourse in machine translation}.

\bibitem[{Sennrich et~al.(2016{\natexlab{a}})Sennrich, Haddow, and
  Birch}]{sennrich-etal-2016-improving}
Rico Sennrich, Barry Haddow, and Alexandra Birch. 2016{\natexlab{a}}.
\newblock \href {https://doi.org/10.18653/v1/P16-1009} {Improving neural
  machine translation models with monolingual data}.
\newblock In \emph{Proceedings of the 54th Annual Meeting of the Association
  for Computational Linguistics (Volume 1: Long Papers)}, pages 86--96, Berlin,
  Germany. Association for Computational Linguistics.

\bibitem[{Sennrich et~al.(2016{\natexlab{b}})Sennrich, Haddow, and
  Birch}]{Sennrich2016BPE}
Rico Sennrich, Barry Haddow, and Alexandra Birch. 2016{\natexlab{b}}.
\newblock \href {https://doi.org/10.18653/v1/P16-1162} {Neural machine
  translation of rare words with subword units}.
\newblock In \emph{Proceedings of the 54th Annual Meeting of the Association
  for Computational Linguistics (Volume 1: Long Papers)}, pages 1715--1725,
  Berlin, Germany. Association for Computational Linguistics.

\bibitem[{Stojanovski and Fraser(2019)}]{Stojanovski2019ImprovingAR}
Dario Stojanovski and Alexander Fraser. 2019.
\newblock \href {https://www.aclweb.org/anthology/W19-6614} {Improving anaphora
  resolution in neural machine translation using curriculum learning}.
\newblock In \emph{Proceedings of Machine Translation Summit XVII Volume 1:
  Research Track}, pages 140--150, Dublin, Ireland. European Association for
  Machine Translation.

\bibitem[{Tiedemann(2012)}]{Europarl}
J{\"o}rg Tiedemann. 2012.
\newblock \href
  {http://www.lrec-conf.org/proceedings/lrec2012/pdf/463_Paper.pdf} {Parallel
  data, tools and interfaces in {OPUS}}.
\newblock In \emph{Proceedings of the Eighth International Conference on
  Language Resources and Evaluation ({LREC}'12)}, pages 2214--2218, Istanbul,
  Turkey. European Language Resources Association (ELRA).

\bibitem[{Tiedemann and Scherrer(2017)}]{tiedemann-scherrer-2017-neural}
J{\"o}rg Tiedemann and Yves Scherrer. 2017.
\newblock \href {https://doi.org/10.18653/v1/W17-4811} {Neural machine
  translation with extended context}.
\newblock In \emph{Proceedings of the Third Workshop on Discourse in Machine
  Translation}, pages 82--92, Copenhagen, Denmark. Association for
  Computational Linguistics.

\bibitem[{Ugawa et~al.(2018)Ugawa, Tamura, Ninomiya, Takamura, and
  Okumura}]{ugawa-etal-2018-neural}
Arata Ugawa, Akihiro Tamura, Takashi Ninomiya, Hiroya Takamura, and Manabu
  Okumura. 2018.
\newblock \href {https://www.aclweb.org/anthology/C18-1274} {Neural machine
  translation incorporating named entity}.
\newblock In \emph{Proceedings of the 27th International Conference on
  Computational Linguistics}, pages 3240--3250, Santa Fe, New Mexico, USA.
  Association for Computational Linguistics.

\bibitem[{Vaswani et~al.(2017)Vaswani, Shazeer, Parmar, Uszkoreit, Jones,
  Gomez, Kaiser, and Polosukhin}]{Vaswani-17-transformer}
Ashish Vaswani, Noam Shazeer, Niki Parmar, Jakob Uszkoreit, Llion Jones,
  Aidan~N Gomez, {\L}ukasz Kaiser, and Illia Polosukhin. 2017.
\newblock Attention is all you need.
\newblock In \emph{Advances in Neural Information Processing Systems 30},
  NIPS~'17, pages 5998--6008.

\bibitem[{Voita et~al.(2019)Voita, Sennrich, and
  Titov}]{Voita2019ContextAwareMR}
Elena Voita, Rico Sennrich, and Ivan Titov. 2019.
\newblock \href {https://doi.org/10.18653/v1/D19-1081} {Context-aware
  monolingual repair for neural machine translation}.
\newblock In \emph{Proceedings of the 2019 Conference on Empirical Methods in
  Natural Language Processing and the 9th International Joint Conference on
  Natural Language Processing (EMNLP-IJCNLP)}, pages 877--886, Hong Kong,
  China. Association for Computational Linguistics.

\bibitem[{Voita et~al.(2018)Voita, Serdyukov, Sennrich, and
  Titov}]{Voita-ACL18}
Elena Voita, Pavel Serdyukov, Rico Sennrich, and Ivan Titov. 2018.
\newblock Context-aware neural machine translation learns anaphora resolution.
\newblock In \emph{Proceedings of the 56th Annual Meeting of the Association
  for Computational Linguistics}, ACL~'18, pages 1264--1274, Melbourne,
  Australia.

\bibitem[{Wang et~al.(2017)Wang, Utiyama, Liu, Chen, and
  Sumita}]{Wang2017InstanceWF}
Rui Wang, Masao Utiyama, Lemao Liu, Kehai Chen, and Eiichiro Sumita. 2017.
\newblock \href {https://doi.org/10.18653/v1/D17-1155} {Instance weighting for
  neural machine translation domain adaptation}.
\newblock In \emph{Proceedings of the 2017 Conference on Empirical Methods in
  Natural Language Processing}, pages 1482--1488, Copenhagen, Denmark.
  Association for Computational Linguistics.

\bibitem[{Wong et~al.(2020)Wong, Maruf, and Haffari}]{Wong2020ContextualNM}
KayYen Wong, Sameen Maruf, and Gholamreza Haffari. 2020.
\newblock \href {https://doi.org/10.18653/v1/2020.acl-main.530} {Contextual
  neural machine translation improves translation of cataphoric pronouns}.
\newblock In \emph{Proceedings of the 58th Annual Meeting of the Association
  for Computational Linguistics}, pages 5971--5978, Online. Association for
  Computational Linguistics.

\bibitem[{Yakhnenko et~al.(2005)Yakhnenko, Silvescu, and
  Honavar}]{yakhnenko2005discriminatively}
Oksana Yakhnenko, Adrian Silvescu, and Vasant Honavar. 2005.
\newblock Discriminatively trained markov model for sequence classification.
\newblock In \emph{Fifth IEEE International Conference on Data Mining
  (ICDM'05)}, pages 8--pp. IEEE.

\bibitem[{Yang et~al.(2019)Yang, Chen, Wang, and Xu}]{Yang2019EffectivelyTN}
Zhen Yang, Wei Chen, Feng Wang, and Bo~Xu. 2019.
\newblock Effectively training neural machine translation models with
  monolingual data.
\newblock \emph{Neurocomputing}, 333:240--247.

\bibitem[{Zhang et~al.(2018)Zhang, Luan, Sun, Zhai, Xu, Zhang, and
  Liu}]{zhang-etal-2018-improving}
Jiacheng Zhang, Huanbo Luan, Maosong Sun, Feifei Zhai, Jingfang Xu, Min Zhang,
  and Yang Liu. 2018.
\newblock \href {https://doi.org/10.18653/v1/D18-1049} {Improving the
  transformer translation model with document-level context}.
\newblock In \emph{Proceedings of the 2018 Conference on Empirical Methods in
  Natural Language Processing}, pages 533--542, Brussels, Belgium. Association
  for Computational Linguistics.

\bibitem[{Zheng et~al.(2020)Zheng, Zhou, Huang, Li, Dai, and
  Chen}]{mirror_Zheng2020Mirror-Generative}
Zaixiang Zheng, Hao Zhou, Shujian Huang, Lei Li, Xin-Yu Dai, and Jiajun Chen.
  2020.
\newblock \href {https://openreview.net/forum?id=HkxQRTNYPH} {Mirror-generative
  neural machine translation}.
\newblock In \emph{International Conference on Learning Representations}.

\end{thebibliography}

\appendix
\section{Appendix}

\begin{table}[h]
\centering
\scalebox{0.71}{\begin{tabular}{lll}
\toprule
\textbf{Model} & \textbf{Paramaters} & \textbf{Values} \\

\midrule

\concat & --optimizer & adam \\
& --adam-betas & `(0.9, 0.98)' \\
& --clip-norm & 0.0 \\
& --lr-scheduler & inverse\_sqrt \\
& --warmup-init-lr & 1e-07 \\
& --warmup-updates & 4000 \\
& --lr & 0.0007 \\
& --min-lr & 1e-09 \\
& --criterion & label\_smoothed\_cross\_entropy \\
& --label-smoothing & 0.1 \\
& --weight-decay & 0.0 \\
& --max-tokens & 4096 \\
& --update-freq & 8 \\
& --share-all-embeddings & - \\
& --max-update & 100000  \\
\hline
\sts & \textit{as in \concat} & \textit{as in \concat} \\

\bottomrule
\end{tabular}}

\caption{Training parameters used for \sts\ and \concat\ models.  }
\label{table:training_parameters}

\end{table}

\subsection{Training Parameters}
\label{subsec:train_parameters}
The training parameters used for both the \sts\ and the \concat\ models are given in Table~\ref{table:training_parameters}. All models were trained in \textbf{fairseq} and all results reported are based on averaging the last 10 checkpoints. 

\subsection{Examples from Fine-tuned Models}
Some examples of improved translations from our fine-tuned models are given in Table~\ref{table:examples}.

\begin{table*}[t!]
\small 
\centering
\scalebox{0.85}{\begin{tabular}{ll}
\toprule
\textbf{Description} & \multicolumn{1}{c}{\textbf{Examples}} \\
\midrule
\multicolumn{2}{c}{\textbf{WMT14 Testset}} \\
\midrule

Source & 14 stunden kämpften die ärzte um das überleben des opfers , jedoch vergeblich .\\
Reference & for 14 hours, doctors battled to save the life of the victim , ultimately in vain .\\
Baseline &  \red{14 hours of doctors} fought for the victim's survival , but in vain .
\\
Our best model & \blue{the doctors fought 14 hours} for the survival of the victim , but in vain . \\
 \midrule
 
Source & der handel am nasdaq options market wurde am freitagnachmittag deutscher zeit unterbrochen . \\

Reference & trading at the nasdaq options market was interrupted on friday afternoon , german time .\\
 
Baseline & trade at nasdaq options market was cut off \red{on the german friday afternoon .}\\

Our best model & trade in nasdaq options market was suspended \blue{on friday afternoon in germany .}\\
 
\midrule
Source & einem autofahrer wurde eine strafe in höhe von 1.000 £ auferlegt , weil er mit bis zu 210 km / h \\
& und einem heißgetränk zwischen seinen beinen gefahren war . \\

Reference & a motorist has been fined £ 1,000 for driving at up to 130mph ( 210km / h ) with a hot drink \\
& balanced between his legs .\\

Baseline & a driver was fined £ 1,000 for driving up to \red{£ 210 per hour} and a hot drink between his legs .\\

Our best model & a driver was fined £ 1,000 for driving up to \blue{210 kilometers an hour} and a hot drink between his legs .\\

\midrule
Source & des grues sont arrivées sur place peu après 10 heures , et la circulation sur la nationale a été détournée dans la foulée . \\

Reference & cranes arrived on the site just after 10am , and traffic on the main road was diverted afterwards . \\

Baseline & cranes arrived soon after 10 hours , and \red{circulation on the national front was hijacked in the process} .\\

Our best model & cranes arrived shortly after 10 hours , and \blue{traffic on the national side was diverted along the way} .\\

\midrule

Source &  le diagnostic de rage a été confirmé par l'institut pasteur .\\

Reference & the diagnosis of rabies was confirmed by the pasteur institute .\\

Baseline & the rabies diagnosis was confirmed by the \red{institut pasteur}.\\ 

Our best model & the rabies diagnosis was confirmed by the \blue{pasteur institute} .\\

\midrule
\multicolumn{2}{c}{\textbf{Pronoun Testset}} \\
\midrule

Context & ... die die amerikanische flamme in die umnachtete welt bringe : lady liberty geht voran .\\
Source & sie soll die fackel der freiheit von den vereinigten staaten in den rest der welt tragen . \\
Context & ... taking the american flame out to the benighted world : \textbf{lady liberty} is stepping forward . \\
Reference & she is meant to be carrying the torch of liberty from the united states to the rest of the world .\\
Baseline & \red{it} is meant to carry the torch of freedom from the united states to the rest of the world . \\
Our best model & \blue{she} is supposed to carry the torch of freedom from the united states to the rest of the world . \\

\midrule
Context & der getestete 1,6 l diesel mit 88 kw / 120 ps beschleunigt den hr - v ...\\
Source & er dürfte seine arbeit allerdings etwas leiser verrichten .\\
Context & the 1.6 l \textbf{diesel engine} we tested , with 88 kw / 120 horsepower accelerates the hr - v powerfully ...\\
Reference & however , it could certainly do its work a bit more quietly .\\
Baseline & however , \red{he} is likely to do \red{his} job rather more quietly .\\
Our best model & but \blue{it} is likely to do \blue{its} job a little more quietly . \\

\midrule

Context & versteinerte reste der haut bedecken noch immer die holprigen panzerplatten , die den schädel des tieres tragen . \\
Source &  sein rechter vorderfuß liegt an seiner seite , seine fünf finger sind nach oben gespreizt . \\
Context & fossilized remnants of skin still cover the bumpy armor plates dotting the \textbf{animal's} skull .\\
Reference &  its right forefoot lies by its side , its five digits splayed upward . \\
Baseline & \red{his} right - hand front foot is on \red{his} side , \red{his} five fingers are spiked up .\\
Our best model & \blue{its} right front foot is on \blue{its} side , \blue{its} five fingers are split upwards . \\

\midrule
Context & Il est mort dimanche matin. \\
Source &  elle avait promis à son mari , la semaine avant son décès , de le faire sortir de l'hôpital \\
Context & \textbf{He} died on Sunday morning. \\

Reference & a week before his death , she had promised her husband she would get him out of hospital \\

Baseline & she promised her husband , the week before \red{she} died , to take \red{her} out of the hospital .\\

Our best model & she promised her husband , the week before \blue{his} death , to take \blue{him} out of the hospital \\

\midrule
Context & Elle a été détenue dans une cellule du commissariat local avant l'audience devant le tribunal. \\
 Source & elle était en vacances dans la région de krabi , au sud de la thaïlande .\\

Context & \textbf{She} was held in local police cells before the court hearing. \\
 Reference & she was holidaying at the resort area of krabi in southern thailand .\\

 Baseline & \red{it} is on holiday in the region of krabi , southern thailand .\\

 Our best model & \blue{she} was on holiday in the krabi region of southern thailand .\\

\bottomrule
\end{tabular}}

\caption{Examples showing the improvements in translations from our best models, across the WMT14 and the pronoun testsets. The previous sentence context information for the pronoun testset is also shown.  }
\label{table:examples}

\end{table*}

\end{document}